\documentclass{article}

\usepackage[preprint]{neurips_2026}

\usepackage[utf8]{inputenc} 
\usepackage[T1]{fontenc}    
\usepackage{hyperref}       
\usepackage{url}            
\usepackage{booktabs}       
\usepackage{amsfonts}       
\usepackage{nicefrac}       
\usepackage{amsmath}
\usepackage{microtype}      
\usepackage{xcolor}         
\usepackage{enumitem}
\usepackage{algorithm}
\usepackage{algpseudocode}
\usepackage{wrapfig}
\renewcommand{\vec}{\boldsymbol}

\usepackage{graphicx}
\usepackage{float}

\usepackage[most]{tcolorbox}

\title{VIMPO: Value-Implicit Policy Optimization for LLMs}

\author{%
Zhewei Kang \\
UC Berkeley\\
{\small \texttt{waynekang07@gmail.com}} \\
\And
Aosong Feng\\
Yale University\\
{\small \texttt{aosong.feng@yale.edu}}\\
\And 
Sergey Levine\\
UC Berkeley\\
{\small \texttt{svlevine@berkeley.edu}}\\
\And
Dawn Song\\
UC Berkeley\\
{\small \texttt{dawnsong@berkeley.edu}}\\
\And
Xuandong Zhao\\
UC Berkeley\\
{\small \texttt{xuandongzhao@berkeley.edu}} \\
}

\begin{document}

\maketitle

\begin{abstract}Reinforcement learning with verifiable rewards has become a central tool for improving the reasoning ability of large language models, but current methods face a trade-off between simplicity and credit assignment. Group-relative methods such as GRPO avoid training a critic, but typically assign a trajectory-level advantage to every token. Actor-critic methods provide denser learning signals, but require a learned value function with its own training instability. We introduce VIMPO, a critic-free policy optimization method that derives a policy-implied value function from the optimality conditions of KL-regularized reinforcement learning. For autoregressive generation, the resulting value recurrence can be written in terms of policy-reference log-ratios and anchored by the terminal condition that no future reward remains at the end of a trajectory. This gives a simple value loss that incorporates outcome-level verifiable rewards without training a critic. The same derivation also yields a critic-free actor advantage, allowing VIMPO to separate reward incorporation through the value loss from policy improvement through a PPO-style actor update. On mathematical RLVR benchmarks, VIMPO improves over GRPO across MATH-500, AIME 2024, AIME 2025, and OlympiadBench, with especially larger gains on competition-style evaluations. Under noisy rewards, VIMPO retains a consistent advantage over GRPO, suggesting that policy-implied value optimization can provide finer credit assignment while preserving the practical simplicity of critic-free training. Code is available
at \url{https://github.com/backprop07/VIMPO}.
\end{abstract}

\section{Introduction}
Reinforcement learning has emerged as the central post-training paradigm for large language models (LLMs), particularly for tasks demanding multi-step reasoning such as mathematical problem solving and code generation~\citep{ouyang2022training,deepseekr1,jaech2024openai}. Existing RL post-training methods for LLMs largely divide into actor-critic and critic-free approaches. Actor-critic methods such as PPO~\citep{schulman2017ppo} and VAPO~\citep{yue2025vapo} retain a learned value function, enabling dense token-level supervision but making training sensitive to critic quality and critic-policy co-adaptation~\citep{shao2025vapo_limits}. In contrast, group-relative methods such as GRPO~\citep{shao2024deepseekmath} and DAPO~\citep{yu2025dapo} eliminate the critic and estimate advantages from groups of rollouts for the same prompt. This simplifies training by replacing the learned state-dependent baseline with a sample-based group baseline, at the cost of broadcasting the same advantage across all tokens. The result is a practical tension between the stability and simplicity of critic-free methods and the dense credit assignment provided by actor-critic methods.

This tension is especially important for long-form reasoning. Group-relative methods are attractive because they avoid training a value network that must remain calibrated to a continually evolving policy. Yet this simplicity comes with a coarse credit assignment signal. A single trajectory-level advantage is typically broadcast across all tokens, treating a completion as an indivisible unit rather than separating pivotal reasoning steps from routine connective tokens. Recent work attempts to recover token-level granularity \emph{post hoc}. FIPO~\citep{qwen2026fipo} re-weights the GRPO advantage using a future-KL factor, while attention-based credit assignment methods~\citep{li2025attention} identify ``anchor'' tokens heuristically. These remedies are useful, but they remain add-ons to objectives that were not designed for fine-grained credit assignment.

We argue that this trade-off is not fundamental. We model autoregressive generation as a deterministic-transition MDP and analyze its KL-regularized policy optimization objective. This yields an identity that represents the value function implicitly through the policy log-likelihood ratio against a frozen reference model. Instead of learning a separate critic, the resulting method estimates the initial value from the same Monte Carlo rollout groups used in Reinforcement Learning with Verifiable Rewards~(RLVR) and anchors the recurrence with the terminal condition that the value at the end of a trajectory is zero. The terminal condition closes the Bellman recurrence, and the Monte Carlo estimate provides the trajectory-level reward target. This gives a critic-free Bellman-consistency objective that aligns the policy-implied value function with outcome-level rewards.

\begin{figure}[t]
    \centering
    \includegraphics[width=0.9\linewidth]{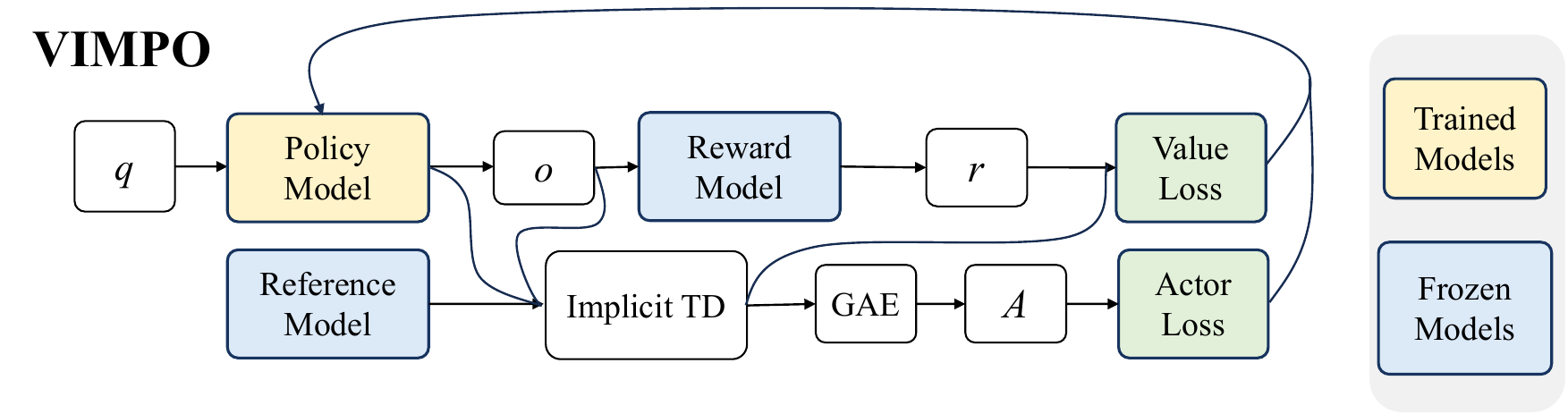}
    \caption{Overview of VIMPO. Given a prompt $q$, the policy generates a
    completion $o$, scored to obtain an outcome reward $r$. VIMPO uses this
    reward to train the policy-implied value loss, while the policy and frozen
    reference model define a token-level TD signal used to form the actor advantage. 
    This separates reward incorporation from policy improvement without
    training an explicit critic.}
    \label{fig:vimpo_overview}
\end{figure}

This construction also yields a closed-form one-step temporal-difference advantage that can be plugged into a PPO-style actor update. The resulting algorithm, \textbf{VIMPO} (\textbf{V}alue-\textbf{IM}plicit \textbf{P}olicy \textbf{O}ptimization), combines critic-free training with Bellman-consistent token-level credit assignment as shown in Figure~\ref{fig:vimpo_overview}. VIMPO therefore occupies a middle ground between group-relative methods and actor-critic methods. It keeps the simplicity of RLVR-style group training while recovering the dense supervision that motivates learned critics. Moreover, because its actor update is expressed through policy-internal log-ratio quantities while its value objective incorporates external verifiable rewards, VIMPO gives a coherent mechanism for connecting outcome-based supervision with internal policy-derived learning signals. Our experiments show that this structure improves training efficiency and reduces sensitivity to noisy rewards. Our contributions are fourfold.
\begin{itemize}[leftmargin=*, itemsep=0pt, topsep=0pt]  
\item We derive a closed-form representation of the optimal value function for a KL-regularized policy on the deterministic-transition MDP induced by autoregressive generation. The representation is expressed through policy log-ratios against a frozen reference model, together with a Monte Carlo estimate of the initial value (Section~\ref{sec:recurrence}).

\item We use the terminal boundary condition $V^*(\vec{s}_T)=0$ to obtain a critic-free value optimization objective. The resulting squared-error loss trains a policy-implied value function without requiring a separately learned critic (Section~\ref{sec:loss}).

\item We show that the same construction yields a closed-form one-step temporal-difference advantage expressed entirely through policy log-ratios. This advantage can be integrated into PPO-style actor updates and extended with GAE-style multi-step aggregation, enabling token-level credit assignment without a learned critic (Section~\ref{sec:ppo_integration}).

\item We empirically compare VIMPO with the critic-free GRPO baseline. With tuned hyperparameters, VIMPO improves faster during training and reaches higher validation accuracy. Under noisy rewards, both methods degrade, but VIMPO is less affected by reward noise (Section~\ref{sec:experiments}).
\end{itemize}

\section{Related Work}
\label{sec:related}

\begin{wrapfigure}{r}{0.53\linewidth}
\vspace{-1em}
    \centering
    \includegraphics[width=0.99\linewidth]{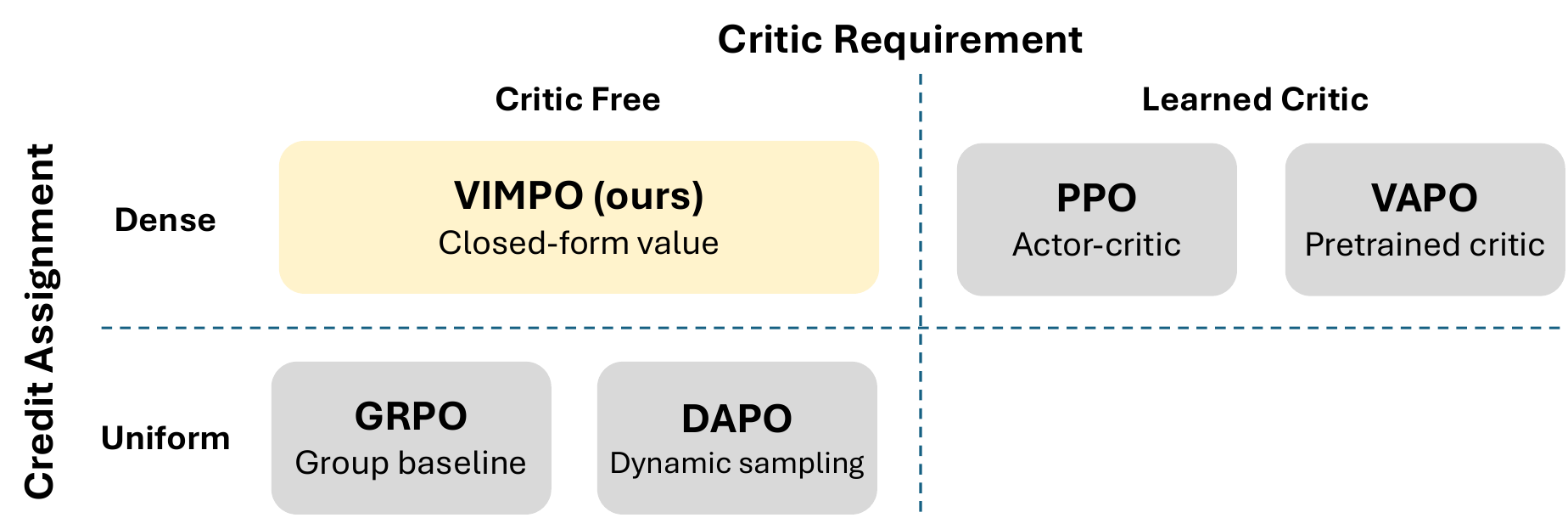}
    \vspace{-1.4em}
    \caption{VIMPO is a critic-free method with token-level credit assignment.}
    \label{fig:main_fig}
    \vspace{-1em}
\end{wrapfigure}
\paragraph{Actor-critic methods with learned values.} PPO~\citep{schulman2017ppo} remains the standard actor-critic algorithm for policy optimization and is commonly paired with generalized advantage estimation~\citep{schulman2015gae}. In LLM post-training, actor-critic variants train a value model alongside the policy to provide dense token-level credit assignment. VAPO~\citep{yue2025vapo} adapts this paradigm to long-CoT reasoning with value pretraining, decoupled GAE, length-adaptive estimation, and token-level policy-gradient updates. These methods provide fine-grained learning signals but require an additional value model, whose errors can affect policy optimization. VIMPO targets the same credit-assignment role without training a separate critic.

\paragraph{Critic-free and post-hoc credit assignment methods.} GRPO~\citep{shao2024deepseekmath} removes the learned value network by estimating advantages from a group of rollouts sampled for the same prompt. DAPO~\citep{yu2025dapo} builds on this critic-free paradigm with dynamic sampling, token-level loss aggregation, and decoupled clipping for long-CoT training. These methods avoid critic pretraining and critic-policy co-training, but their advantage signal is usually derived at the trajectory level and shared across tokens. Recent work therefore attempts to refine this coarse signal post hoc. FIPO~\citep{qwen2026fipo} reweights the group-relative advantage using a future-KL factor, while attention-based credit assignment methods~\citep{li2025attention} identify important tokens from internal attention statistics. These approaches introduce token-level structure through auxiliary weighting rules. VIMPO instead derives its token-level actor signal from the KL-regularized optimality conditions and a policy-implied value recurrence.

\paragraph{Log-ratio parameterizations and Bellman consistency.} DPO~\citep{rafailov2023dpo} shows that the optimal policy of a KL-regularized preference objective can be expressed through a log-ratio against a reference policy, avoiding explicit reward-model training. IPO~\citep{azar2024ipo} further analyzes preference optimization under related log-ratio parameterizations, and TDPO~\citep{zeng2024tdpo} introduces token-wise log-ratio terms while remaining in the offline pairwise-preference setting. VIMPO also uses policy-reference log-ratios but applies them to online RL with rollout rewards. The Bellman side of our derivation is related to classical consistency-based RL. Path Consistency Learning~\citep{nachum2017pcl} trains policies and values by enforcing multi-step consistency conditions, while soft Q-learning~\citep{haarnoja2018sql} derives policy updates from entropy-regularized Bellman equations. VIMPO shares the idea of deriving learning signals from Bellman-style identities, but uses the standard expected-return value in a KL-regularized LLM post-training objective. This keeps the reference-policy log-ratio explicit and yields a critic-free value and actor update for autoregressive generation.

\section{Methodology}
\label{sec:method}

We derive VIMPO in four steps. Section~\ref{sec:setup} formalizes autoregressive generation as a deterministic-transition MDP and states the KL-regularized policy optimization problem. Section~\ref{sec:recurrence} derives a value recurrence from the KL-regularized optimality condition and the Bellman equation. Section~\ref{sec:loss} converts this recurrence into a critic-free value optimization objective using a terminal boundary condition. Section~\ref{sec:ppo_integration} uses the resulting policy-implied value function to construct a critic-free PPO-style actor update.

\subsection{Setup: autoregressive generation as a deterministic MDP}
\label{sec:setup}

For a prompt $\vec{x} \in \mathcal{V}^{\ell_x}$, an autoregressive policy $\pi$ generates a completion $a_0, a_1, \ldots, a_{T-1}$, where each $a_t \in \mathcal{V}$. We define the token-prefix state as
\[
\vec{s}_t = (\vec{x}, a_0, \ldots, a_{t-1}) \in \mathcal{V}^{\ell_x+t}, \qquad t \in [0,T],
\]
with $\vec{s}_0=\vec{x}$ and the terminal state $\vec{s}_T$. Given the next token $a_t$, the transition is deterministic:
\[
P(\vec{s}_{t+1}\mid \vec{s}_t,a_t)=1.
\]

For a fixed policy $\pi$, we use the standard finite-horizon expected-return action-value function:
\[
Q^\pi(\vec{s}_t,a_t) = \mathbb{E}_{\pi} \left[ \sum_{k=t}^{T-1} \gamma^{k-t} r(\vec{s}_k,a_k) \;\middle|\; \vec{s}_t,a_t \right],
\]
and the corresponding value function:
\[
V^\pi(\vec{s}_t) = \mathbb{E}_{a\sim\pi(\cdot\mid\vec{s}_t)} \left[ Q^\pi(\vec{s}_t,a) \right].
\]
Our value function is the standard expected-return value used in actor-critic RLHF, rather than the entropy-regularized soft value used in Path Consistency Learning and soft Q-learning~\citep{nachum2017pcl,haarnoja2018sql}. As a result, we can control how KL enters the value and actor objectives, and we avoid requiring estimates of future KL contributions when constructing the advantage. This separation is consistent with recent RLVR practice, where reducing or removing explicit KL penalties can improve reasoning performance in some settings~\citep{yu2025dapo}.

At optimality, the KL-regularized policy satisfies the local optimality condition:
\begin{equation}
\pi^*(\cdot \mid \vec{s}_t) = \arg\max_{\pi} \; \mathbb{E}_{a \sim \pi(\cdot \mid \vec{s}_t)} \left[ Q^*(\vec{s}_t,a) \right] - \beta\, \mathrm{KL} \left( \pi(\cdot \mid \vec{s}_t) \,\|\, \pi_{\mathrm{ref}}(\cdot \mid \vec{s}_t) \right),
\label{eq:obj}
\end{equation}
where \(Q^*\) is the optimal action-value function associated with the same fixed point. Solving the local maximization gives:
\begin{equation}
\pi^*(a \mid \vec{s}_t) = \frac{ \pi_{\mathrm{ref}}(a \mid \vec{s}_t) }{ Z(\vec{s}_t) } \exp \left( \frac{1}{\beta} Q^*(\vec{s}_t,a) \right).
\label{eq:softmax}
\end{equation}

\subsection{From KL-regularized optimality to value recurrence}
\label{sec:recurrence}

The deterministic transition collapses the Bellman equation to:
\begin{equation}
Q^*(\vec{s}_t, a_t) = r(\vec{s}_t, a_t) + \gamma V^*(\vec{s}_{t+1}).
\label{eq:bellman}
\end{equation}
Combining \eqref{eq:bellman} with \eqref{eq:softmax}, and eliminating  $\ln Z$ via the identity  $\ln Z(\vec{s}_t) = \tfrac{1}{\beta} V^*(\vec{s}_t) -  \mathrm{KL}^*(\vec{s}_t)$, where  $\mathrm{KL}^*(\vec{s}_t) := \mathrm{KL}(\pi^*(\cdot\mid\vec{s}_t)\| \pi_{\mathrm{ref}}(\cdot\mid\vec{s}_t))$, yields the central identity of this paper:
\begin{equation}
\beta \ln \frac{\pi^*(a_t \mid \vec{s}_t)}{\pi_{\mathrm{ref}}(a_t \mid \vec{s}_t)}  \;=\; r(\vec{s}_t, a_t) + \gamma V^*(\vec{s}_{t+1}) - V^*(\vec{s}_t) \;+\; \beta \,\mathrm{KL}^*(\vec{s}_t).
\label{eq:identity}
\end{equation}

Equation~\eqref{eq:identity} is the backbone of VIMPO. It states that, under optimality, the per-token log-ratio against the reference policy \emph{is} a Bellman residual plus a KL correction. The following three observations immediately follow:
\begin{itemize}[leftmargin=*, itemsep=0pt, topsep=0pt]  
\item \textbf{Closed-form value.} Rearranging \eqref{eq:identity} as $\gamma V^*(\vec{s}_{t+1}) - V^*(\vec{s}_t) = \beta\ln(\pi^*/\pi_{\mathrm{ref}}) - \beta\,\mathrm{KL}^* - r$ defines a one-step recurrence for $V^*$. Given any anchor value $V^*(\vec{s}_0)$, we can solve this recurrence forward without ever evaluating a learned critic.
\item \textbf{Zero-mean signal.} The bracket $\beta\ln(\pi^*/\pi_{\mathrm{ref}}) - \beta\,\mathrm{KL}^*$ has expectation zero under $a_t \sim \pi^*(\cdot\mid\vec{s}_t)$, since $\mathrm{KL}^*$ is exactly the expectation of the log-ratio. This property, derived rather than imposed, will give VIMPO its training stability.
\item \textbf{Closed-form one-step advantage.} The Bellman-residual form of \eqref{eq:identity} gives a closed-form one-step TD advantage  $A^{\mathrm{TD}}_t = r + \gamma V^* - V^* =  \beta\ln(\pi^*/\pi_{\mathrm{ref}}) - \beta\,\mathrm{KL}^*$, which we exploit in Section~\ref{sec:ppo_integration} for PPO integration.
\end{itemize}

\subsection{Training the policy-implied value}
\label{sec:loss}

Let $V_t := V^*(\vec{s}_t)$ and $B_t := \beta\ln(\pi^*(a_t\mid\vec{s}_t)/ \pi_{\mathrm{ref}}(a_t\mid\vec{s}_t)) - \beta\,\mathrm{KL}^*(\vec{s}_t) - r(\vec{s}_t, a_t)$. Equation~\eqref{eq:identity} is equivalent to $B_t = \gamma V_{t+1} - V_t$, a first-order linear recurrence. Solving forward from $V_0$ yields
\begin{equation}
V^*(\vec{s}_t) = \gamma^{-t} V^*(\vec{s}_0) + \sum_{k=0}^{t-1} \gamma^{k-t} B_k.
\label{eq:recurrence}
\end{equation}

\paragraph{Constructing the implied value.}
Equation~\eqref{eq:recurrence} characterizes the optimal value function. To turn this into a training objective, we follow the DPO~\citep{rafailov2023dpo} strategy of substituting the trainable policy $\pi$ for $\pi^*$ in the closed-form expression. This defines the \emph{policy-implied value}:
\begin{equation}
V_\pi(\vec{s}_t) := \gamma^{-t} \widehat{V}_0 +  \sum_{k=0}^{t-1} \gamma^{k-t}  \!\left[ \beta\ln\frac{\pi(a_k\mid\vec{s}_k)}{\pi_{\mathrm{ref}}(a_k\mid\vec{s}_k)}  - \beta\,\mathrm{KL}_\pi(\vec{s}_k) - r(\vec{s}_k, a_k) \right],
\label{eq:vpi}
\end{equation}
where $\widehat{V}_0$ is a Monte Carlo estimate of the initial value and $\mathrm{KL}_\pi(\vec{s}_k) := \mathrm{KL}(\pi(\cdot\mid\vec{s}_k)\|\pi_{\mathrm{ref}}(\cdot\mid\vec{s}_k))$.

\paragraph{The terminal anchor.}
At the terminal state, no future reward remains, so $V_\pi(\vec{s}_T) = 0$. This gives a single, parameter-free supervision target, since at optimality the policy-implied value must vanish at the end of every trajectory. We train $\pi$ to satisfy this condition:
\begin{equation}
\mathcal{L}_V(\pi) := \tfrac{1}{2}\,V_\pi(\vec{s}_T)^2.
\label{eq:lossV}
\end{equation}

\paragraph{Specialization to outcome-only rewards.}
For the common LLM RL setting where $r(\vec{s}_k, a_k) = 0$ for $k < T-1$ and $r(\vec{s}_{T-1}, a_{T-1}) = R_{\mathrm{final}}$, with $\gamma = 1$ and $\widehat{V}_0$ estimated by the group-mean return $\overline{R}_{\mathrm{final}}$, equation~\eqref{eq:lossV} unfolds into
\begin{equation}
\mathcal{L}_V(\pi) = \frac{1}{2}\left[ \sum_{k=0}^{T-1}\!\left( \beta\ln\frac{\pi(a_k\mid\vec{s}_k)}{\pi_{\mathrm{ref}}(a_k\mid\vec{s}_k)}  - \beta\,\operatorname{sg}\!\left[\mathrm{KL}_\pi(\vec{s}_k)\right] \right) - \big(R_{\mathrm{final}} - \overline{R}_{\mathrm{final}}\big) \right]^2
\label{eq:lossfinal}
\end{equation}
where $\operatorname{sg}[\cdot]$ denotes the stop-gradient operator. This is the operational form of the value-loss component of VIMPO. The target is centered by the group-mean reward, matching the normalization used in group-relative RLVR methods. We stop gradients through the KL term so that it acts only as a centering baseline. Otherwise, the value loss could be reduced by directly manipulating the full-distribution KL term rather than by aligning sampled-token log-ratios with centered rollout rewards. Intuitively, the loss encourages the cumulative policy-reference log-ratio along a trajectory to track the centered final reward, tying the policy-implied value function to rollout outcomes without introducing a learned critic.

\begin{algorithm}[t]
\caption{VIMPO training step}
\label{alg:vimpo}
\begin{algorithmic}[1]
\Require Policy model $\pi_\theta$, reference model $\pi_{\mathrm{ref}}$, prompt distribution $\mathcal{D}$, group size $G$, coefficients $\beta,c_A$, learning rate $\eta$
\For{each training step}
    \State Sample prompt $\vec{x}\sim\mathcal{D}$ and rollouts $\{\tau^{(i)}\}_{i=1}^G \sim \pi_\theta(\cdot\mid\vec{x})$
    \State Compute final rewards $R^{(i)}$ and group baseline
    $\overline{R}=\frac{1}{G}\sum_i R^{(i)}$
    \For{each rollout $i$ and token $t$}
        \State Compute
        $\rho^{(i)}_t=\beta\log \frac{\pi_\theta(a^{(i)}_t\mid \vec{s}^{(i)}_t)}{\pi_{\mathrm{ref}}(a^{(i)}_t\mid \vec{s}^{(i)}_t)}$
        \State Compute $\kappa^{(i)}_t=\beta\,\mathrm{KL}\bigl(\pi_\theta(\cdot\mid\vec{s}^{(i)}_t) \| \pi_{\mathrm{ref}}(\cdot\mid\vec{s}^{(i)}_t)\bigr)$
        \State Set policy-implied advantage $A^{(i)}_t=\rho^{(i)}_t-\kappa^{(i)}_t$
    \EndFor
    \State Compute value loss $\mathcal{L}_V=\frac{1}{2G}\sum_i \left[\sum_t(\rho^{(i)}_t-\operatorname{sg}[\kappa^{(i)}_t]) -(R^{(i)}-\overline{R})\right]^2$
    \State Normalize detached actor advantages over valid response tokens:
    $\widetilde{A}^{(i)}_t =
    \frac{A^{(i)}_t-\mu_{\mathcal{B}}}{\sigma_{\mathcal{B}}+\epsilon}$
    \State Compute PPO actor loss $\mathcal{L}_A=\mathrm{PPO\text{-}Clip} \bigl(\operatorname{sg}[\{\widetilde{A}^{(i)}_t\}],\pi_\theta,\pi_{\theta_{\mathrm{old}}}\bigr)$
    \State Update $\theta\leftarrow\theta-\eta\nabla_\theta(\mathcal{L}_V+c_A\mathcal{L}_A)$
\EndFor
\end{algorithmic}
\end{algorithm}

\subsection{PPO integration via the closed-form advantage}
\label{sec:ppo_integration}

The same identity also induces a one-step advantage estimator. For a trainable policy $\pi_\theta$, we define the policy-implied TD advantage
\begin{equation}
\widehat{A}^{\mathrm{TD}}_t := \beta\log\frac{\pi_\theta(a_t\mid\vec{s}_t)} {\pi_{\mathrm{ref}}(a_t\mid\vec{s}_t)} - \beta\,\mathrm{KL}_{\pi_\theta}(\vec{s}_t).
\label{eq:advantage}
\end{equation}
This quantity corresponds to the Bellman residual $r(\vec{s}_t,a_t)+\gamma V_{\pi_\theta}(\vec{s}_{t+1})-V_{\pi_\theta}(\vec{s}_t)$ when the policy-implied value satisfies the recurrence in Equation~\ref{eq:vpi}. In practice, we treat $\widehat{A}^{\mathrm{TD}}_t$ as an advantage estimate for a PPO-style actor update.

The one-step estimates can be accumulated in the same form as GAE:
\begin{equation}
\widehat{A}^{\lambda}_t = \sum_{\ell=0}^{T-t-1} (\gamma\lambda)^\ell \widehat{A}^{\mathrm{TD}}_{t+\ell}.
\label{eq: vimpo_gae}
\end{equation}

As is common in PPO implementations, before applying the PPO surrogate we normalize the detached advantages over valid response tokens in each optimization step:
\begin{equation}
\widetilde{A}^{\lambda}_t
=
\frac{
\widehat{A}^{\lambda}_t-\mu_{\mathcal B}
}{
\sigma_{\mathcal B}+\epsilon
},
\end{equation}
where $\mu_{\mathcal B}$ and $\sigma_{\mathcal B}$ are computed over valid response tokens in the minibatch. We then use $\widetilde{A}^{\lambda}_t$ in the PPO actor loss $\mathcal{L}_A$, with no gradient propagated through the advantage normalization..

The external reward enters VIMPO through the value-loss term, not directly through the actor advantage in Equation~\ref{eq:advantage}. Thus, the value loss incorporates outcome-level reward information, while the actor loss uses the induced log-ratio advantage for policy improvement. This separates reward incorporation from policy improvement and links verifiable rewards to policy-internal learning signals.

Formally, let $\mathcal{L}_A$ denote the negative clipped PPO surrogate, and $c_A \geq 0$ control the actor update strength. The overall VIMPO training objective is then:
\begin{equation}
\mathcal{L}_{\mathrm{VIMPO}}(\pi) = \mathcal{L}_V(\pi) + c_A\,\mathcal{L}_A(\pi).
\label{eq:final}
\end{equation}

In implementation, each training step samples groups of rollouts for prompts in the batch, computes the centered final reward, evaluates the policy-reference log-ratio terms along each trajectory, forms the value loss in Equation~\ref{eq:lossfinal}, and applies the PPO actor loss in Equation~\ref{eq:final}. Algorithm~\ref{alg:vimpo} summarizes the resulting training step.

\section{Experiments}
\label{sec:experiments}

We evaluate VIMPO against GRPO in an RLVR setting for mathematical reasoning. We first train both methods on a math-reasoning dataset for an extended number of steps and compare their validation performance and learning dynamics. We then stress-test both methods under noisy rewards to evaluate robustness to imperfect reward signals. Finally, we conduct ablations over key hyperparameters and monitor additional training metrics to better understand the behavior of VIMPO.

\subsection{Experiments setup}
\label{sec:experiment}

We use the math subset of Guru~\citep{cheng2025revisitingreinforcementlearningllm} as the RLVR training set. Guru provides a larger curated collection of verifiable math problems than DAPO-Math~\citep{yu2025dapo}, allowing us to study training behavior over longer runs. Unless otherwise stated, all runs start from Qwen3-4B-Base~\citep{yang2025qwen3} and are implemented in verl~\citep{sheng2024hybridflow}.

We evaluate on MATH-500~\citep{hendrycks2021math}, AIME 2024~\citep{maa2024aime}, AIME 2025~\citep{maa2025aime}, and OlympiadBench~\citep{he2024olympiadbench}. For MATH-500 and AIME 2024, we use the same prompt format and veRL evaluation pipeline used during training. For AIME 2025 and OlympiadBench, we use the LightEval implementation~\citep{lighteval2025}. Because these evaluation pipelines differ slightly in prompting and answer extraction, we compare methods only within the same benchmark.

We compare VIMPO with GRPO under the same shared hyperparameters, listed in Appendix~\ref{app:hparams}. For the main comparison, we use  $\beta=5\times10^{-4}$ and actor coefficient $c_A=5\times10^{-3}$ for VIMPO. These values are selected by coarse tuning  on training dynamics and then held fixed for the main runs. We do not tune shared hyperparameters, such as the learning rate, separately for each method, so that differences in performance primarily reflect the optimization objective rather than method-specific tuning. 

For the GRPO baseline, we use token-level loss aggregation and disable the KL
reward penalty, following common RLVR practice and the DAPO training
recipe~\citep{yu2025dapo}. We also include a naive GRPO baseline, which uses
the original sequence-mean-token-mean GRPO loss~\citep{shao2024deepseekmath}. We do not include additional DAPO-style improvements such as dynamic sampling or filtering. This keeps the comparison focused on the effect of replacing the group-relative objective with the VIMPO value loss and actor update.

For the noisy-reward stress test, we corrupt the training reward by flipping the binary correctness signal independently with probability $25\%$ within each prompt group. The corrupted rewards are used only for optimization. Training accuracy and validation accuracy are always reported using the clean, uncorrupted verifier.

\subsection{Main comparison}
\label{sec:main_comparison}

\begin{figure}[t]
    \centering
    \includegraphics[width=0.99\linewidth]{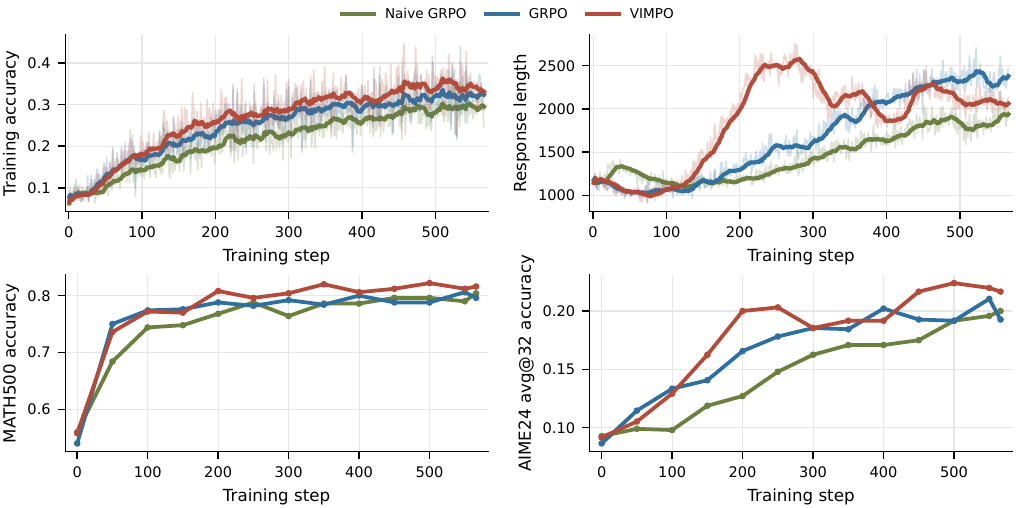}
    \caption{Main comparison among naive GRPO, GRPO and VIMPO under clean verifier rewards. We report training accuracy, response length, MATH-500 accuracy, and AIME 2024 avg@32 accuracy over training. Faint curves show raw logged values, while solid curves show a 15-step moving average for visualization.}
    \label{fig:main_training}
\end{figure}

We first compare VIMPO and the two GRPO baselines under clean golden rewards. Figure~\ref{fig:main_training} reports training accuracy, response length, MATH-500 accuracy, and AIME 2024 avg@32 accuracy during RL fine-tuning. All RL
methods improve substantially over the base model, but VIMPO shows stronger
learning dynamics than both GRPO baselines in the later stage of training and
achieves the highest validation accuracy on both MATH-500 and AIME 2024.

Table~\ref{tab:final_eval} gives the final clean-reward evaluation results. All
RL methods improve substantially over Qwen3-4B-Base, confirming that the
training setup provides a strong RLVR signal. The token-level GRPO baseline
outperforms naive GRPO on the overall average, consistent with prior findings
that token-level loss aggregation is a stronger implementation choice for GRPO~\citep{yu2025dapo}.
VIMPO further improves over both GRPO variants, achieving the best average
performance and the highest score on each benchmark. The improvement over GRPO
is especially pronounced on AIME 2025, where VIMPO improves from $17.6$ to
$20.8$, suggesting that the policy-implied value loss and actor update are most
helpful on harder competition-style reasoning evaluations.

These results suggest that the policy-implied value loss and critic-free actor update provide a useful learning signal beyond the group-relative baseline. The response-length curve in Figure~\ref{fig:main_training} also indicates that VIMPO's gains are not simply explained by monotonically increasing completion length. VIMPO explores longer responses early in training, but its response length later decreases while validation accuracy remains higher than GRPO.

\begin{table}[t]
\centering
\caption{Final evaluation accuracy. Clean rows are evaluated after full training,
while noisy rows are evaluated after 200 training steps with $25\%$ reward
flipping. Avg. is the unweighted average over the four benchmark columns.}
\label{tab:final_eval}
\begin{tabular}{lccccc}
\toprule
\textbf{Method} & \textbf{MATH-500} & \textbf{AIME24} & \textbf{AIME25} & \textbf{OlympiadBench} & \textbf{Avg.} \\
& \textbf{avg@1} & \textbf{avg@32} & \textbf{avg@32} & \textbf{avg@1} &  \\
\midrule
Qwen3-4B-Base & 54.0 & 8.6 & 3.6 & 18.0 & 21.1 \\
\midrule
Naive GRPO & 80.4 & 20.0 & 14.6 & 31.3 & 36.6 \\
GRPO    & 79.6 & 19.3 & 17.6 & 33.2 & 37.4 \\
VIMPO  & \textbf{81.6} & \textbf{21.7} & \textbf{20.8} & \textbf{34.0} & \textbf{39.5} \\
\midrule
GRPO noisy    & 75.2 & 13.0 & 11.7 & 28.9 & 32.2 \\
VIMPO noisy   & \textbf{78.2} & \textbf{18.3} & \textbf{14.9} & \textbf{32.3} & \textbf{35.9} \\
\bottomrule
\end{tabular}
\end{table}


\subsection{Robustness to noisy rewards}
\label{sec:noisy_rewards}

\begin{figure}[tb]
    \centering
    \includegraphics[width=0.99\linewidth]{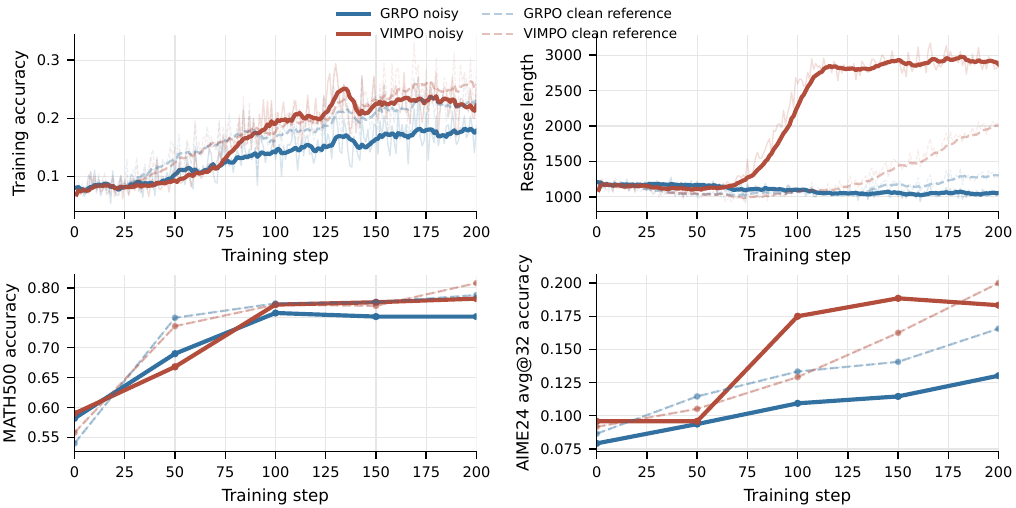}
    \caption{Noisy-reward stress test over the first 200 training steps. Solid curves show runs trained with $25\%$ reward flipping, while faint dashed curves show clean-reward runs for reference. Metrics are evaluated with the clean verifier.}
    \label{fig:noisy_reward}
\end{figure}

Figure~\ref{fig:noisy_reward} shows the noisy-reward stress test described in Section~\ref{sec:experiment}. We plot the first 200 training steps under $25\%$ reward flipping, with clean-reward runs shown as faint references.

The lower two rows of Table~\ref{tab:final_eval} show that reward noise degrades both methods, but VIMPO retains a consistent advantage across all benchmarks. The gap is especially large on the AIME evaluations, suggesting that VIMPO's update is less sensitive to corrupted reward labels in the competition-style reasoning setting.

The training curves in Figure~\ref{fig:noisy_reward} are consistent with this interpretation. Under corrupted rewards, VIMPO improves faster in training accuracy and maintains stronger AIME 2024 validation performance during the early training window. This behavior suggests a possible advantage of separating reward incorporation from policy improvement. In VIMPO, the external reward enters through the value-loss target, while the actor update uses a policy-implied log-ratio advantage. The actor update is therefore not directly proportional to each individual reward label, which may reduce sensitivity to corrupted rewards compared with GRPO's group-relative advantage. This property is useful in practical RLVR settings, where verifiers and reward rules can be imperfect.

\subsection{Ablations}
\label{sec:ablations}

\begin{figure}[t]
    \centering
    \includegraphics[width=0.9\linewidth]{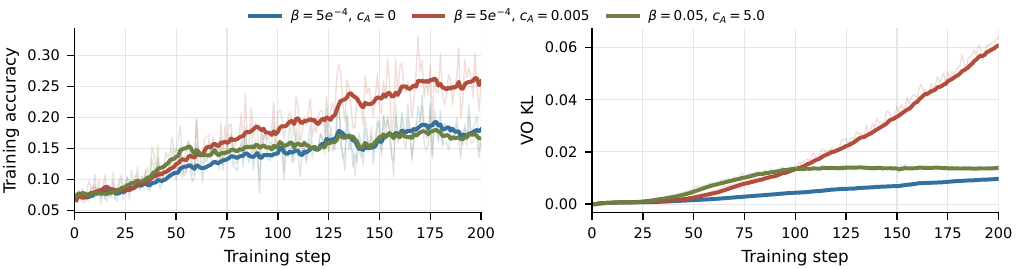}
    \caption{Ablation over VIMPO coefficients. We compare a value-only variant with $c_A=0$, the main VIMPO setting, and a high-$\beta$, high-$c_A$ setting. The actor update improves training accuracy but also increases VO KL, while the high-$\beta$, high-$c_A$ setting keeps KL controlled at the cost of slower learning.}
    \label{fig:beta_factor_ablation}
\end{figure}

We ablate the two VIMPO-specific coefficients, the value coefficient $\beta$ and the actor coefficient $c_A$. Figure~\ref{fig:beta_factor_ablation} shows training accuracy and VO KL over the first 200 training steps for three representative settings. The value-only variant with $c_A=0$ improves over training, but learns more slowly than the full VIMPO objective. Adding the PPO-style actor update with $c_A=0.005$ substantially accelerates learning and achieves the highest training accuracy among the tested settings.

The ablation also shows that this gain comes with a larger policy shift. The main VIMPO setting reaches a much higher VO KL than the value-only variant, indicating that the actor update enables faster movement away from the reference policy. In contrast, the high-$\beta$, high-$c_A$ setting keeps VO KL nearly flat after the early stage, but its accuracy remains close to the value-only baseline. This suggests that overly strong value scaling and actor weighting can over-constrain the effective update and reduce the benefit of the actor component.

Interestingly, the high-$\beta$, high-$c_A$ setting improves quickly in the first 50 steps before plateauing. This suggests that stronger early regularization or actor weighting may help initial optimization, but can become limiting later in training. A natural extension is to adapt or anneal $\beta$ and $c_A$ over training, using larger values early for stabilization and smaller values later to permit further policy improvement.

Overall, the ablation supports the design of VIMPO as a combination of the value loss and actor update. The value loss alone provides a usable learning signal, while the actor update is important for faster optimization. At the same time, the coefficients must be chosen carefully, since stronger policy updates can affect response-length dynamics, which we report in Appendix~\ref{app:response_length_ablation}.

\section{Discussion and Limitations}
\label{sec:discussion}

\paragraph{Fixed reference and coefficient schedules.}
One limitation of the current implementation is that we keep $\beta$ fixed throughout training and use a frozen reference policy. The usefulness of this reference-based constraint may change as the policy moves farther from initialization. In particular, a fixed $\beta$ may overemphasize the reference-policy geometry in later training, when further performance improvements may require moving beyond the region where the reference provides a useful local guide. This concern is consistent with recent RLVR practice, where reducing or removing explicit KL penalties can improve reasoning performance in some settings~\citep{yu2025dapo}. Our ablations further show that the actor weight $c_A$ also affects learning speed, policy movement, and response-length dynamics. A natural extension is to adapt $\beta$ and $c_A$, or periodically update the reference policy, so that the objective provides early stabilization without imposing the same reference-based geometry throughout optimization.

\paragraph{Cost of exact KL.}
Our implementation uses exact full-distribution KL to match the derivation, but this requires access to the full next-token distributions of both the policy and reference model. For larger models or longer responses, this can become a nontrivial overhead. Approximate KL estimators, candidate-set approximations, or cached reference distributions may improve scalability, but they may also change the centering properties used by the value objective.

\paragraph{Experimental scope.}
Our experiments focus on mathematical RLVR with a 4B-scale base model and use GRPO as the primary critic-free baseline. Due to the computational cost of long-horizon RLVR training, we report single-seed experiments and restrict the main comparison to this model scale and domain. This setting provides clean verifiable rewards and a controlled comparison, but it does not establish that the same gains transfer unchanged to code, tool use, open-ended instruction following, robotics, or larger model scales. We also do not compare against a carefully tuned actor-critic baseline such as PPO or VAPO. Such a comparison would be valuable, especially in domains where dense state-dependent value estimates may be more important than in outcome-only math RLVR. Evaluating VIMPO against tuned actor-critic methods across additional verifiable domains, random seeds, and model sizes is an important direction for future work.

\section{Conclusion}
\label{sec:conclusion}

We introduced VIMPO, a critic-free policy optimization method for RLVR that derives value and actor signals from the optimality conditions of KL-regularized reinforcement learning. By modeling autoregressive generation as a deterministic-transition process, VIMPO expresses the value recurrence in terms of policy-reference log-ratios and uses a terminal boundary condition to train a policy-implied value function without a learned critic. The same construction yields a critic-free actor advantage that can be used in a PPO-style update.

Empirically, VIMPO improves over GRPO on mathematical reasoning benchmarks, with stronger gains on competition-style evaluations and better robustness under noisy rewards. These results suggest that policy-implied value optimization can recover some of the fine-grained learning signal associated with actor-critic methods while preserving the simplicity of critic-free RLVR. Future work should study adaptive coefficient schedules, approximate KL estimators, reference-policy updates, and extensions beyond mathematical reasoning to other verifiable domains.

\bibliographystyle{plainnat}
\bibliography{references}

\newpage

\appendix

\section{Experimental Details}
\label{app:experimental_details}

\subsection{Shared hyperparameters}
\label{app:hparams}

Table~\ref{tab:shared_hparams} lists the hyperparameters shared by GRPO and
VIMPO in the main experiments. Method-specific VIMPO coefficients are reported
in Section~\ref{sec:experiment}.

\begin{table}[H]
\centering
\caption{Shared experimental hyperparameters for GRPO and VIMPO.}
\label{tab:shared_hparams}
\begin{tabular}{ll}
\toprule
\textbf{Hyperparameter} & \textbf{Value} \\
\midrule
Base model & Qwen3-4B-Base \\
Training data & Guru Math subset, 54.4K examples \\
Max prompt length & 2048 \\
Max response length & 4096 \\
Responses per prompt & 8 \\
Prompt batch size & 96 \\
Generated responses per batch & 768 \\
Rollout temperature & 1.0 \\
Rollout top-$p$ & 1.0 \\
Validation temperature & 1.0 \\
Validation top-$p$ & 0.7 \\
Mini-batch size & 96 \\
Actor learning rate & $1\times 10^{-6}$ \\
Warmup steps & 10 \\
Weight decay & 0.1 \\
Gradient clipping & 1.0 \\
\bottomrule
\end{tabular}
\end{table}

\subsection{GRPO baseline loss aggregation}
\label{app:grpo_loss}

For the GRPO baselines, we keep the advantage estimator group-relative and
consider two loss aggregations. The main GRPO baseline uses token-level loss
aggregation following the DAPO training recipe~\citep{yu2025dapo}. We also
include a naive GRPO baseline using the original sequence-mean-token-mean
aggregation~\citep{shao2024deepseekmath}. Both are different from the VIMPO
advantage in Equation~\ref{eq:advantage}. For a
prompt $\vec{x}$, we sample a group of completions
$\{\tau^{(i)}\}_{i=1}^{G}$ from $\pi_{\theta_{\mathrm{old}}}$, where
$\tau^{(i)}=(a^{(i)}_0,\ldots,a^{(i)}_{T_i-1})$. Let $R^{(i)}$ be the final
reward for completion $\tau^{(i)}$. The group-normalized GRPO advantage is
\begin{equation}
\widehat{A}^{\mathrm{GRPO}}_i
=
\frac{
R^{(i)}-\overline{R}
}{
\operatorname{std}\!\left(\{R^{(j)}\}_{j=1}^{G}\right)+\epsilon
},
\qquad
\overline{R}
=
\frac{1}{G}\sum_{j=1}^{G}R^{(j)} .
\label{eq:grpo_advantage}
\end{equation}
The same scalar $\widehat{A}^{\mathrm{GRPO}}_i$ is assigned to every valid
token in completion $\tau^{(i)}$.

For each token, define the PPO probability ratio and clipped token objective as
\begin{align}
r^{(i)}_t(\theta)
&=
\frac{
\pi_\theta(a^{(i)}_t \mid \vec{s}^{(i)}_t)
}{
\pi_{\theta_{\mathrm{old}}}(a^{(i)}_t \mid \vec{s}^{(i)}_t)
},
\label{eq:grpo_ratio}
\\
\ell^{(i)}_t(\theta)
&=
\min\Bigg(
r^{(i)}_t(\theta)\widehat{A}^{\mathrm{GRPO}}_i,\,
\operatorname{clip}\!\left(
r^{(i)}_t(\theta),
1-\epsilon_{\mathrm{low}},
1+\epsilon_{\mathrm{high}}
\right)
\widehat{A}^{\mathrm{GRPO}}_i
\Bigg).
\label{eq:grpo_token_objective}
\end{align}
The token-mean clipped GRPO objective used for the main GRPO baseline is
\begin{equation}
\mathcal{J}_{\mathrm{GRPO}}
=
\mathbb{E}_{\vec{x}\sim\mathcal{D}}
\mathbb{E}_{\{\tau^{(i)}\}_{i=1}^{G}\sim
\pi_{\theta_{\mathrm{old}}}(\cdot\mid\vec{x})}
\left[
\frac{1}{\sum_{i=1}^{G}T_i}
\sum_{i=1}^{G}\sum_{t=0}^{T_i-1}
\ell^{(i)}_t(\theta)
\right].
\label{eq:grpo_token_mean}
\end{equation}

For the naive GRPO baseline, we use the
sequence-mean-token-mean aggregation:
\begin{equation}
\mathcal{J}_{\mathrm{GRPO}\text{-}\mathrm{seq}}
=
\mathbb{E}_{\vec{x}\sim\mathcal{D}}
\mathbb{E}_{\{\tau^{(i)}\}_{i=1}^{G}\sim
\pi_{\theta_{\mathrm{old}}}(\cdot\mid\vec{x})}
\left[
\frac{1}{G}
\sum_{i=1}^{G}
\frac{1}{T_i}
\sum_{t=0}^{T_i-1}
\ell^{(i)}_t(\theta)
\right].
\label{eq:grpo_sequence_mean}
\end{equation}

In implementation, both objectives are minimized by negating the corresponding
surrogate objective. 
The main GRPO baseline and VIMPO both use token-level loss aggregation, but
they differ in the advantage signal. GRPO broadcasts the group-relative
trajectory advantage $\widehat{A}^{\mathrm{GRPO}}_i$ to all tokens in a
completion, while VIMPO uses the policy-implied token-level advantage in
Equation~\ref{eq:advantage}. The naive GRPO baseline uses the same
group-relative advantage as GRPO, but averages the token loss within each
completion before averaging across completions.

\section{Additional Experimental Results}
\label{app:additional_results}

\subsection{Training entropy}
\label{app:training_entropy}

\begin{figure}[H]
\centering
\includegraphics[width=0.75\linewidth]{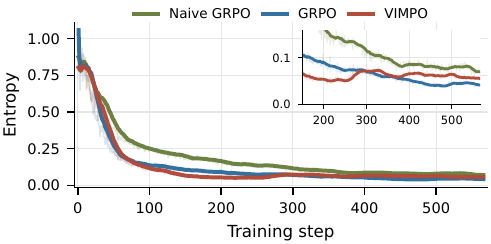}
\caption{Training entropy of naive GRPO, GRPO, and VIMPO under clean
verifier rewards. Faint curves show raw logged values, while solid curves
show a moving average. Both GRPO baselines exhibit a mostly monotonic
decrease in entropy, whereas VIMPO shows small fluctuations after entering
the low-entropy regime.}
\label{fig:training_entropy}
\end{figure}

Figure~\ref{fig:training_entropy} reports the average token-level policy
entropy during training. For a generated trajectory
$\tau=(a_0,\ldots,a_{T-1})$, we compute
\begin{equation}
\mathcal{H}(\pi_\theta) = 
\frac{1}{T}
\sum_{t=0}^{T-1}
\left[
-\sum_{a_t\in\mathcal{V}}
\pi_\theta(a_t\mid \vec{s}_t)
\log \pi_\theta(a_t\mid \vec{s}_t)
\right],
\label{eq:training_entropy}
\end{equation}
and average this quantity over sampled completions.

The two GRPO baselines show a largely monotonic entropy decrease throughout
training. Naive GRPO maintains higher entropy than token-level GRPO for much of
training, but both baselines continue moving toward lower-entropy policies.
VIMPO also rapidly enters a low-entropy regime, but its entropy does not
decrease as monotonically and shows small recoveries in the later stage of
training. This is consistent with the response-length dynamics in
Figure~\ref{fig:main_training}, where VIMPO changes its generation behavior
during training rather than simply increasing confidence monotonically.

These results provide an additional diagnostic difference between VIMPO and
group-relative baselines. The monotonic entropy decrease observed for both GRPO
variants is consistent with the entropy-collapse behavior discussed by
\citet{cui2025entropy}, and may be connected to the uniform token-level
advantage assignment used by group-relative methods. In contrast, VIMPO uses a
policy-implied token-level advantage, which may allow more local adjustment of
the policy distribution after entering the low-entropy regime. We view
Figure~\ref{fig:training_entropy} as suggestive evidence that VIMPO is less
tightly coupled to monotonic entropy collapse, while leaving a controlled study
of this mechanism to future work.

\subsection{Response length in the coefficient ablation}
\label{app:response_length_ablation}

\begin{figure}[H]
    \centering
    \includegraphics[width=0.75\linewidth]{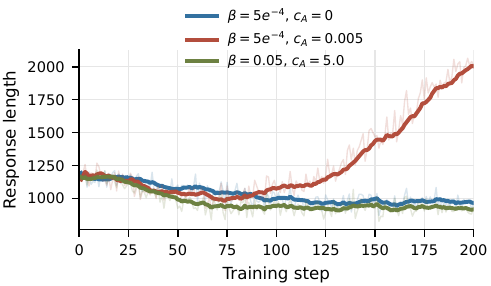}
    \caption{Response-length dynamics for the VIMPO coefficient ablation. The
    main VIMPO setting improves training accuracy fastest, but also produces
    longer responses by step 200.}
    \label{fig:response_length_ablation}
\end{figure}

Figure~\ref{fig:response_length_ablation} reports the response-length dynamics
for the coefficient ablation in Section~\ref{sec:ablations}. The main VIMPO
setting, which uses both the value loss and actor update, learns fastest in
training accuracy but also produces longer responses by step 200. This supports
the observation that VIMPO-specific coefficients affect not only optimization
speed and policy movement, but also generation-length dynamics.

\section{Qualitative Token-Level Signal Visualization}
\label{app:token_visualization}

We provide a qualitative case study illustrating the token-level signal induced by
VIMPO on a MATH-500 combinatorics problem. Both completions are sampled from a
Qwen3-4B-Base model trained with VIMPO. For the same prompt, the model produces
one correct response and one incorrect response, allowing us to compare the
resulting token-level signals on two closely related reasoning trajectories.

\begin{tcolorbox}[title=Case study: Mr. Potato Head counting problem]
\textbf{Question.}
Tom got a Mr. Potato Head for his birthday. It came with 3 hairstyles, 2 sets of
eyebrows, 1 pair of googly eyes, 2 sets of ears, and 2 sets of lips, a pair of
regular shoes, and a bonus pair of clown shoes. If a complete Mr. Potato Head
personality includes eyebrows, eyes, ears, lips, shoes and optionally hair, how
many different wacky personalities can Tom come up with? Note that Mr. Potato
Head can be bald. You cannot mix and match components within a pair or set.

\textbf{Correct reasoning branch.}
The correct completion treats each pair or set as an indivisible component
choice. Thus the valid counts are
\[
\text{eyebrows}=2,\quad
\text{eyes}=1,\quad
\text{ears}=2,\quad
\text{lips}=2,\quad
\text{shoes}=2,\quad
\text{hair}=4,
\]
where the four hair options are the three hairstyles plus bald. This gives
\[
2 \times 1 \times 2 \times 2 \times 2 \times 4 = 64.
\]

\textbf{Incorrect reasoning branch.}
The incorrect completion treats ``both sets'' as an additional valid option for
some paired components. This changes the counts for eyebrows, ears, and lips
from \(2\) to \(3\), producing
\[
3 \times 1 \times 3 \times 3 \times 2 \times 4 = 216.
\]
\end{tcolorbox}

\begin{figure}[H]
    \centering
    \includegraphics[width=0.95\linewidth]{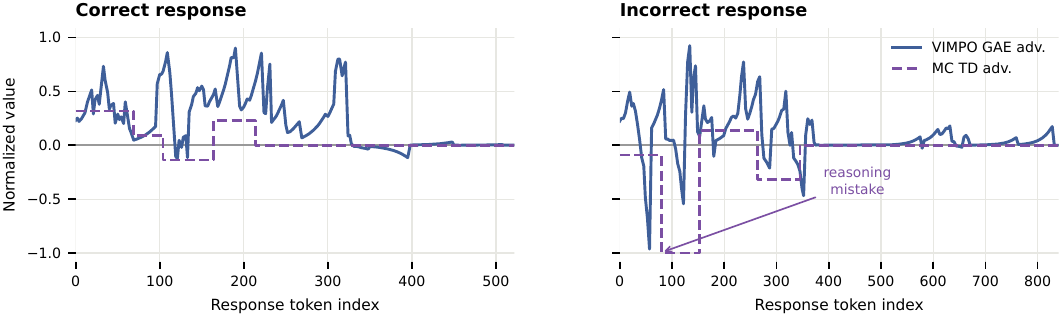}
    \caption{Token-aligned comparison between the VIMPO GAE actor signal and a
    Monte Carlo rollout-based temporal-difference advantage estimate.
    Both curves are normalized using min--max normalization.}
    \label{fig:vimpo_advantage_methods}
\end{figure}

Figure~\ref{fig:vimpo_advantage_methods} compares the VIMPO actor signal with a
Monte Carlo rollout-based diagnostic on one correct and one incorrect completion
for this problem. The VIMPO curve corresponds to the policy-implied GAE
advantage defined in Equation~\ref{eq: vimpo_gae}. The Monte Carlo diagnostic is
computed at 10 selected prefixes by sampling 32 continuations from the current
policy at each prefix, then measuring the change in empirical correctness across
adjacent prefixes to obtain a temporal-difference proxy. Since this diagnostic is
available only at a small number of prefixes, we expand it into a token-level
step function for visual comparison. Both curves are min--max normalized for
visualization.

The Monte Carlo diagnostic confirms that the incorrect trajectory loses
downstream correctness probability after the overcounting branch is introduced.
VIMPO provides a finer-grained token-level signal without requiring additional
rollouts from intermediate prefixes. In the correct completion, the signal is
favorable around the valid decomposition into component counts and the final
multiplication to \(64\). In the incorrect completion, the signal highlights the
local branch that introduces the invalid ``both sets'' option and propagates it
into factors of \(3\), leading to the final answer \(216\).

\begin{figure}[H]
    \centering
    \includegraphics[width=0.95\linewidth]{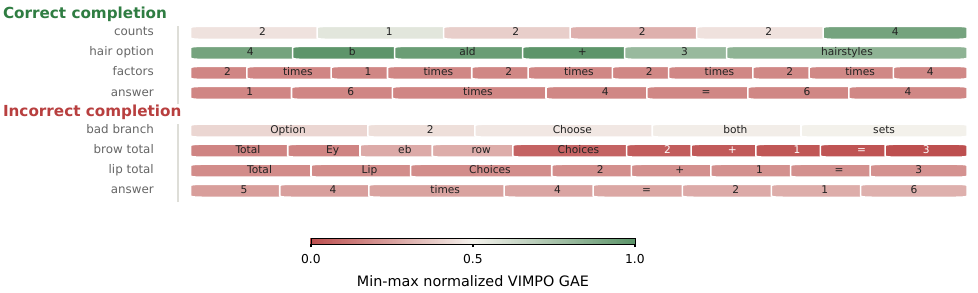}
    \caption{Token-level view of selected critical spans from the same
    combinatorics case using the VIMPO GAE signal. Each cell corresponds to one
    model token, with whitespace and formatting-only tokens omitted for readability.
    Colors indicate normalized VIMPO advantages, highlighting favorable signal on
    the valid counting trajectory and adverse signal near the overcounting branch.}
    \label{fig:vimpo_advantage_tokens}
\end{figure}

Figure~\ref{fig:vimpo_advantage_tokens} shows a token-level view of selected
critical spans from the same example. Each displayed cell corresponds to a model
token, with whitespace and formatting-only tokens omitted for readability. The
colors show the min--max normalized VIMPO GAE signal over the displayed tokens.
The correct span assigns favorable signal to the valid component-level counting
structure, while the incorrect span assigns adverse signal near the overcounting
steps. This illustrates how VIMPO can expose localized reasoning decisions.

Overall, this case study provides a qualitative sanity check for the VIMPO
advantage signal. The coarse Monte Carlo estimate exhibits a broadly consistent
pattern with the policy-implied VIMPO advantage, suggesting that the VIMPO signal
captures meaningful token-level structure in the reasoning trajectory.

\end{document}